# It is not all downhill from here: Syllable Contact Law in Persian

Afshin Rahimi, Moharram Eslami, Bahram Vazirnezhad

## Abstract

Syllable contact pairs cross-linguistically tend to have a falling sonority slope, a constraint which is called the Syllable Contact Law (SCL). In this study, the phonotactics of syllable contacts in 4202 CVC.CVC words of Persian lexicon is investigated. The consonants of Persian were divided into five sonority categories and the frequency of all possible sonority slopes is computed both in lexicon (type frequency) and in corpus (token frequency). Since an unmarked phonological structure has been shown to diachronically become more frequent we expect to see the same pattern for syllable contact pairs with falling sonority slope. The correlation of sonority categories of the two consonants in a syllable contact pair is measured using Pointwise Mutual Information (PMI) both in lexicon level and in corpus level. Results show that SCL is not a categorical constraint in Persian and all possible sonority slopes are observed. In addition evidence show that at lexical level, the less sonority slope (-4 to +4), the more frequent. The reason of frequency increase is shown to be the tendency of non-sonorants such as stops and fricatives to occur in onset position, their reluctance to occur in coda position and the tendency of sonorants such as nasals and liquids to occur in coda position rather than onset position. PMI between sonority categories of two consonants in a syllable contact pair provides evidence against SCL in Persian. In other words, the sonority categories don’t impose any restriction on each other and are not correlated. Higher frequencies of syllable contact pairs with falling sonority slope is not an effect of SCL but the effect of the constraints in coda position of the first syllable and the constraints in onset position of the second syllable.

## 1. Introduction

Syllable Contact Law (SCL) is the tendency of syllable contact pairs to have a falling sonority slope. Based on the assumption that unmarked sequences will diachronically become more frequent, we investigate the frequency of syllable contacts with all possible sonority slopes and found that syllable contact pairs with falling sonority are more frequent in FLexicon, the Persian lexicon corpus. The distribution of sonority categories in onset and coda position of syllable contact pairs is also probed. We argue that higher frequency of syllable contact pairs with falling sonority slope is the result of the tendency of sonorants to be in coda position and the tendency of non-sonorants to be in onset position. We strengthen our argument by providing statistical evidence that the sonority of consonants in a syllable contact pair are not correlated using Pointwise Mutual Information. We have also repeated our experiments in corpus level under the assumption that SCL constraints may show their gradient effects in corpus level based on usage frequency but the same results were obtained. In both lexicon level and corpus level no gradient reflexes of SCL were found.

### 1.1. Sonority

Although sonority has been widely used to explain phonotactics and especially syllable structure, its nature is still controversial. From an articulatory view point sonority indicates extra openness of the mouth. From an auditory point of view it shows higher loudness and it acoustically indicates higher acoustic signal intensity. There are arguments against defining sonority as an inherent property of each segment. According to these arguments sonority of each voice implies its relative loudness comparing with other voices which yields a hierarchy of segments (Ladefoged and Johnson 2011). A hierarchy of segments can be found in (1) based on their relative sonority level.

(1) *Vowels > Liquids > Nasals > Fricatives > Affricates > Stops*

In some sources (Clements 1990; Parker 2002) sonority has been suggested as a group of acoustic features. In some other sources, sonority is argued to be the carrier of linguistic message but not the message itself and so sonority is not like distinctive features (Harris 2005). For a comprehensive review of literature on sonority see Parker (2002, 2004).

### 1.2. Syllable Contact Law

According to the work by Murray and Vennemann (1983) and Vennemann (1988) there is a cross-linguistically preference for syllable contacts with falling sonority slope. This preference is called Syllable Contact Law (SCL). A rewording of SCL is brought here:

> *For all syllable contacts A.B, the more sonority falls from A to B, the more A.B is preferred.*

As an example, everything else being equal /al.ta/ is preferred to /at.la/ because the sonority slopes falls more from /l/ to /t/ than from /t/ to /l/. Languages differ in the amount of sonority rise they allow to be surfaced. To explain this difference among languages of the world Gouskova (2004) argues that Syllable Contact Law is not a single constraint but a relational hierarchy of ranked constraints from the most marked (*Dis+7) to least marked (*Dis-7). Gouskova has used 7 sonority level which results in 15 different sonority slopes from -7 to 7. She argues that the constraints are relational because the ranking of constraints is not dependent on the sonority level of coda or onset alone but on the relation (difference) between the two sonority levels. Figure 1 from Gouskova (2004) shows the amount of sonority rise a language permits using her relational hierarchy of constraints.

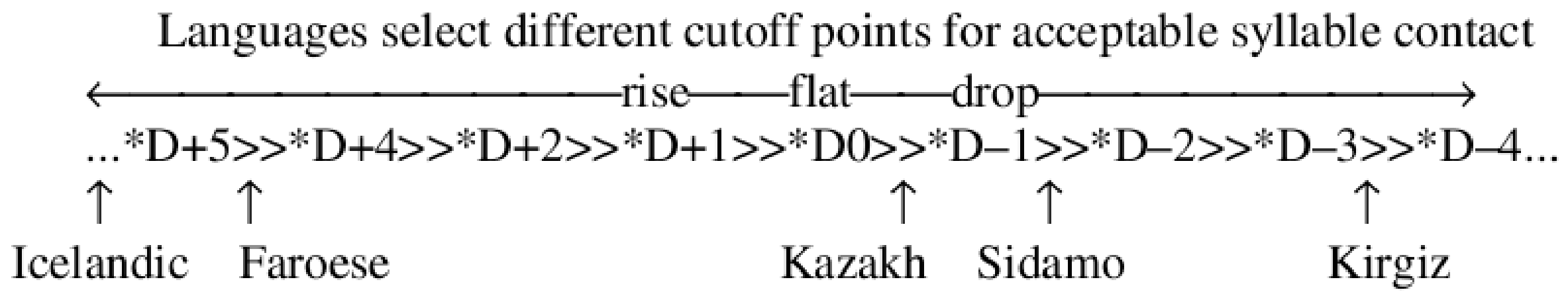

**Figure 1 – Relational hierarchy of SCL constraints and the cutoff points acceptable by each language proposed in Gouskova (2004)**

As it is shown in Figure 1 in some languages like Kirgiz, Sidamo, Kazakh and Faroese SCL constraints have a categorical role. For example in Kirgiz syllable contacts with sonority slope -2 are not attested. Baertsch and Davis (2009) propose another hierarchy of constraints using split margin approach to syllable arguing that the conjunction of marginal constraints in coda and onset positions provides a better model for observed phonotactics in syllable contacts (SCL).

### 1.3. Gradient Lexical Reflexes of Syllable Contact Law

If SCL is active in the brain of speakers and listeners of a language they will show grammatical acceptability levels for syllable contacts which are closely related to the slope and well-formedness of the stimuli. The more falling the sonority slope is, the more grammatical it should be judged. On the other hand there are a lot of literature on the relation between grammatically judgment and lexical frequency (Coleman and Pierrehumbert 1997; Hay, Pierrehumbert and Beckman 2004; Bybee 2001; Coetzee and Pater 2005). So an unmarked syllable contact according to SCL should both be more accepted in grammatical judgment tasks and be more frequent in lexicon. The gradient lexical reflexes of SCL in English are studied by McGowan (2008).

### 1.4. Repair Strategies for Highly Marked Syllable Contacts

In a majority of syllable contacts which are more marked according to SCL (for instance sonority slope of +4), phonological changes such as vowel epenthesis, omission, assimilation and metathesis are applied as repair strategies and produce more unmarked syllable contacts. Table 3 includes some examples of these attested repair strategies in CVC.CVC words in Persian.

| Persian Word | English Translation | Phonemic Transcription | Sonority Slope | Repair Strategi | Surface Form | New Sonority Slope |
|---|---|---|---|---|---|---|
| اعلام | declaration | ? ɜ?.lɑm | +4 | Omission | ? ɜ.lɑm | +1 |
| زودرس | Early | zud.ræs | +4 | Assimilation | zur.ræs | 0 |
| کبریت | matches | kɜb.rit | +4 | Metathesis | kɜr.bit | -4 |
| دادیار | prosecutor | dɑd.yɑr | +4 | Vowel Epenthesis | dɑ.de.yɑr | +1 |

**Table 3 – Repair strategies in Persian for highly marked syllable contacts of CVC.CVC lexemes according to Syllable Contact Law (SCL)**

## 2. Data and methodology of the present research

Using Flexicon, the Persian lexicon (Eslami and coworkers 2006) which includes more than 54000 Persian lexemes, the expected CVC.CVC segment sequences were extracted by means of the phonemic transcriptions. Since syllabic structure of Persian forbids consonant clusters in onset position and also it is necessary for a syllable to have an obligatory consonant at the onset of the syllable, syllabification is simple and deterministic. Each syllable begins with the first consonant before the vowel and it continues to the first consonant before the next vowel. For instance, assuming the phonemic chain of /CVCCVCVCCVCV/ and according to syllabic structure of Persian, its deterministic syllabification shall be in the form of /CVC.CV.CVC.CV.CV/. Words with CVC.CVC structure were extracted and the sonority distance between the first syllable's coda and the second syllable's onset was calculated. The CVCC.CVC structures were not chosen in order to omit the impact of C1 in CVC1C2.C3VC on the following C2C3 syllable contact.

The sonority hierarchy is assumed as below and each one of Persian consonants is categorized in one of the five categories of table 4. As we have defined five sonority levels among consonants, sonority distance between two consonants can be variable between +4 to -4. Distance of +4 means rising sonority distance between two consonants in the boundary of syllable which is the most marked structure from view point of Syllable Contact Law. A word with such a sonority distance has violated falling sonority pattern in the boundary of the syllable. Samples of the whole types of sonority distance could be found in Persian which means Persian doesn't have a cut-off point for sonority slope.

| **Sonority Category** | **Symbol** | **Sonority** | **Num. of Consonants** | **Consonants** |
|---|---|---|---|---|
| Liquids | LI | 5 | 3 | [y, r, l] |
| Nasals | NA | 4 | 3 | [m, n] |
| Fricatives | FR | 3 | 8 | [v, z, ʒ, f, s, ʃ, h, x] |
| Affricates | AF | 2 | 2 | [tʃ, dʒ] |
| Stops | PL | 1 | 8 | [b, d, g, q, ?, p, t, k] |

Table 4 – Sonority hierarchy of Persian consonants used to investigate sonority slopes of syllable contacts.

Words with CVC.CVC were extracted from Persian lexicon and their sonority slope was measured. All possible syllable contact sonority slopes and an example word is shown in table 5. For instance, the word / dʒæmʃid / with syllable contact /m.ʃ/ has sonority slope -1 resulted from subtracting the sonority of FR (the sonority category of / ʃ /) which is 3 by the sonority of NA (the sonority category of / m /) which is 4. All other sonority slopes are calculated the same.

| Persian Word | Transcripted Word | Coda Sonority Category | Onset Sonority Category | Sonority Slope |
|---|---|---|---|---|
| دلدار | dɜl.dɑr | LI | PL | -4 |
| دیرجوش | dir.dʒuʃ | LI | AF | -3 |
| لمبان | lom.bɑn | NA | PL | -3 |
| پرسود | por.sud | LI | FR | -2 |
| پنچر | pæn.tʃær | NA | AF | -2 |
| پوستین | pus.tin | FR | PL | -2 |
| مورمور | mur.mur | LI | NA | -1 |
| جمشید | dʒæm.ʃid | NA | FR | -1 |
| تهچین | tah.tʃin | FR | AF | -1 |
| گچبر | gætʃ.bor | AF | PL | -1 |
| گلریز | gol.riz | LI | LI | 0 |
| سجاد | sædʒ.dʒɑd | AF | AF | 0 |
| افشین | ? æf.ʃin | FR | FR | 0 |
| بیمناک | bim.nɑk | NA | NA | 0 |
| دیدگاه | did.gɑh | PL | PL | 0 |
| گاومیش | gɑv.miʃ | FR | NA | 1 |
| همراه | hæm.rɑh | NA | LI | 1 |
| مجذور | mædʒ.zur | AF | FR | 1 |
| زودجوش | zud.dʒuʃ | PL | AF | 1 |
| ریشریش | riʃ.riʃ | FR | LI | 2 |
| آچمز | ? ɑtʃ.mæz | AF | NA | 2 |
| لطفا | lot.fæn | PL | FR | 2 |
| جاجرود | dʒɑdʒ.rud | AF | LI | 3 |
| نیکنام | nik.nɑm | PL | NA | 3 |
| تدریس | tædris | PL | LI | 4 |

**Table 5 – Sample of all attested sonority category combinations in coda and onset of syllable contact pairs in Persian lexicon**

## 2.1. PMI Pointwise Mutual Information[1]

In order to study the sonority slope in the boundary of syllables, Pointwise Mutual Information or PMI (Church and Hanks 1989) has been used. This criterion indicates how much two events tend to co-occur. Point wise Mutual Information of two events x and y of random variables X and Y is calculated using the following formula.

$$PMI(x, y) = \log_2 \frac{p(x, y)}{p(x) * p(y)}$$

In this formula, p(x, y) is the probability that two events x and y co-occur. p(x) and p(y) are the probability of occurrence of events x and y respectively. So PMI quantifies discrepancy between co-occurrence of the events given the joint probability and the probability of the two events given the individual probabilities, assuming independence of the two variables. The more PMI of the two events, the more tendency of them to co-occur. The negative amount of this measure shows reluctance of the two events to co-occur.

The PMI is used to analyze the tendency of various sonority categories presented in Table 5 to co-occur in syllable contacts. In this analysis the first random variable is assumed the occurrence of a sonority category in the coda position of first syllable and the second random variable has been assumed the occurrence of a sonority class in onset position of the second syllable. For instance consider the event that /t/ occurs in coda position of the first syllable and the event that /l/ occur in onset position of the second syllable. We wish to obtain the tendency of co-occurrence of the two mentioned phonemes. The tendency for co-occurrence will be high if a great deal of the two phonemes' occurrences are happened at the same time that is each of the two phonemes

[1] PMI Pointwise Mutual Information : $PMI(x, y) = \log_2 \frac{p(x,y)}{p(x)*p(y)}$

may exist in coda or onset positions in great quantities, nevertheless co-occurrence of the two events is low. In addition it is possible that the two events rarely exist in the said positions, but they occur together in the majority cases. PMI measure only indicates the tendency for co-occurrence using omission of separate occurrence impact. To further clarify the notion of PMI an example is presented here. Assume that there are 1000 words with CVC.CVC form in FLexicon. Now we wish to obtain the tendency for co-occurrence of phoneme /t/ in coda position of the first syllable and phoneme /l/ in the onset position of the second syllable. In the first case consider that the phoneme /t/ has appeared 50 times in coda position of the first syllable. In addition phoneme /l/ has appeared 25 times in onset position of the second syllable. Therefore occurrence probability of phoneme /t/ in coda position of the first syllable shall be 0.05 and occurrence probability of phoneme /l/ in coda position of the second syllable shall be 0.025. If the two phonemes occur simultaneously in five syllable contacts as /t.l/, their PMI is obtained using the following formula:

$$PMI(t,l) = \log_2 \frac{p(t,l)}{p(t) * p(l)} = \log_2 \frac{\frac{5}{1000}}{\frac{50}{1000} * \frac{25}{1000}} = 2$$

In this formula, p(t,l) is 0.005, p(t) is 0.05 and p(l) is 0.025. Hence, the amount of PMI will be 2. Positive amount of PMI shows the tendency of phonemes /t/ and /l/ to occur simultaneously in mentioned positions in syllable contact.

In the second case, If we assume that the phoneme /t/ is used 800 times in coda position of the first syllable and phoneme /l/ is used 400 times in onset position of the second syllable in CVC.CVC sequences and additionally the number of co-occurrences of these variables in the syllable contacts is 20, the amount of PMI must be obtained using the same method.

$$PMI(t,l) = \log_2 \frac{p(t,l)}{p(t) * p(l)} = \log_2 \frac{\frac{20}{1000}}{\frac{800}{1000} * \frac{400}{1000}} = -4$$

The negative amount of PMI in the second case indicates that despite the two above phonemes occur more frequently in onset and coda positions of syllable contacts and their co-occurrence is even higher compared to the first case, since their tendency to occur simultaneously with other phonemes is more common, the amount of their PMI shall be negative.

### 2.2. Extraction of frequencies and calculation of PMI

For each consonant cluster C1.C2 in the syllable contacts, sonority slope is obtained by subtracting the sonority of C1 from sonority of C2. The frequency of each syllable contact is also counted both in lexicon (type frequency) and in corpus (token frequency).

We consider both type and token frequencies. The assumption behind this consideration is that a restriction may apply either in lexical level or in usage level. Therefore the gradual effects of the restriction may not be available in lexical level but appear in the corpus level so that more unmarked lexemes may be used more frequently in day to day usage of the language.

## 3. Results

The relation between sonority slope of syllable contacts and their type frequency is shown in chart 1. The red trend line is the overall tendency for changes, obtained by interpolation. A gradual decrease in the number of syllable contacts can be clearly viewed from falling sonority slopes (negative slopes) toward rising sonority slopes (positive slopes).

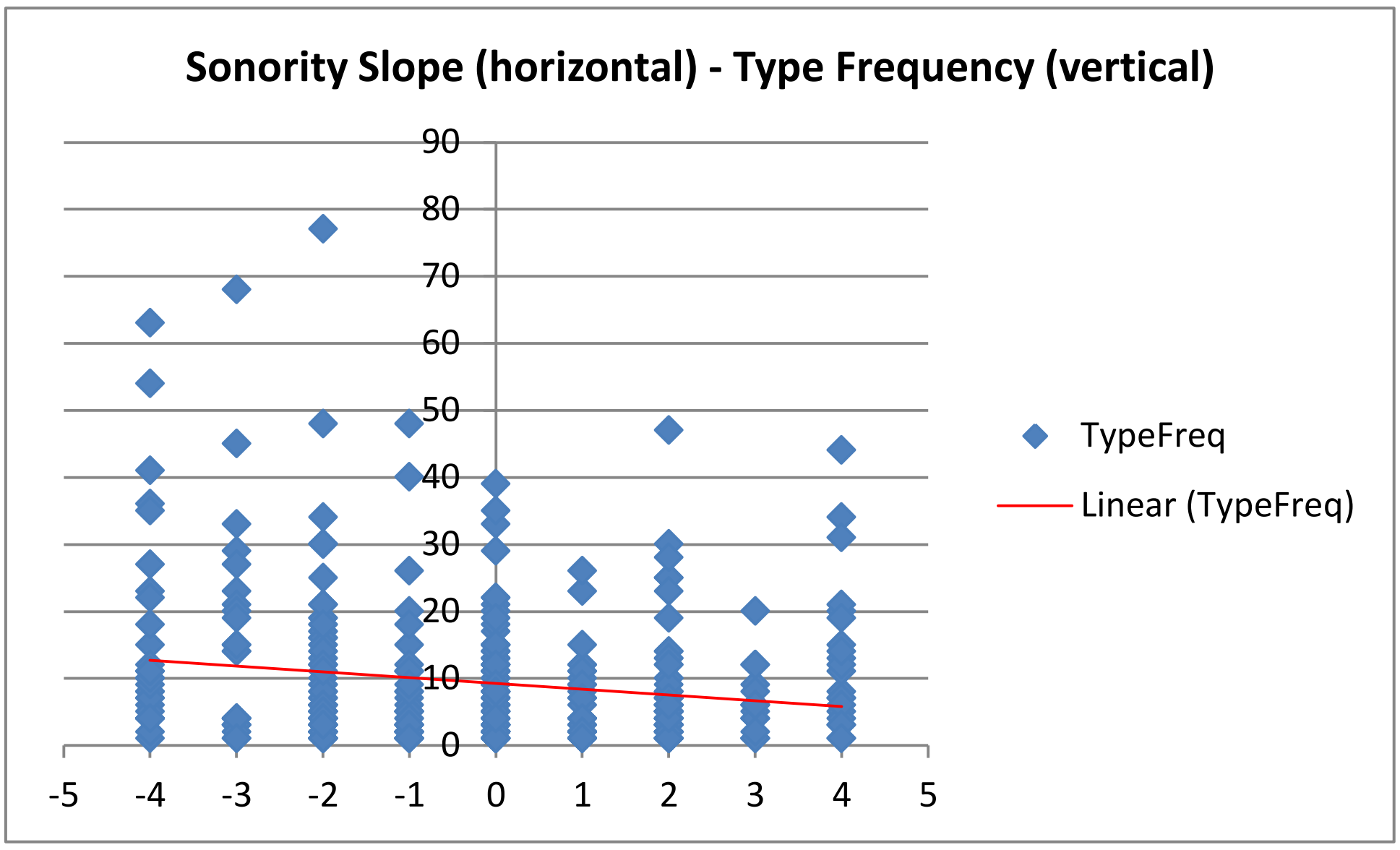

**Diagram 1 – Type frequency of sonority slope in syllable contact pairs**

In chart 2, occurrence probability of each sonority category in onset and coda positions has been compared by using token frequency of syllable contacts. As it can be seen, the stop and affricate sonority categories are more probable to occur in onset position than coda position. This clearly shows why syllable contacts with falling sonority slope are of a high frequency according to chart 1. More sonorant categories tend to exist in coda position and non-sonorant categories tend to occur in onset position and therefore structures with falling sonority slope are more probable to occur.

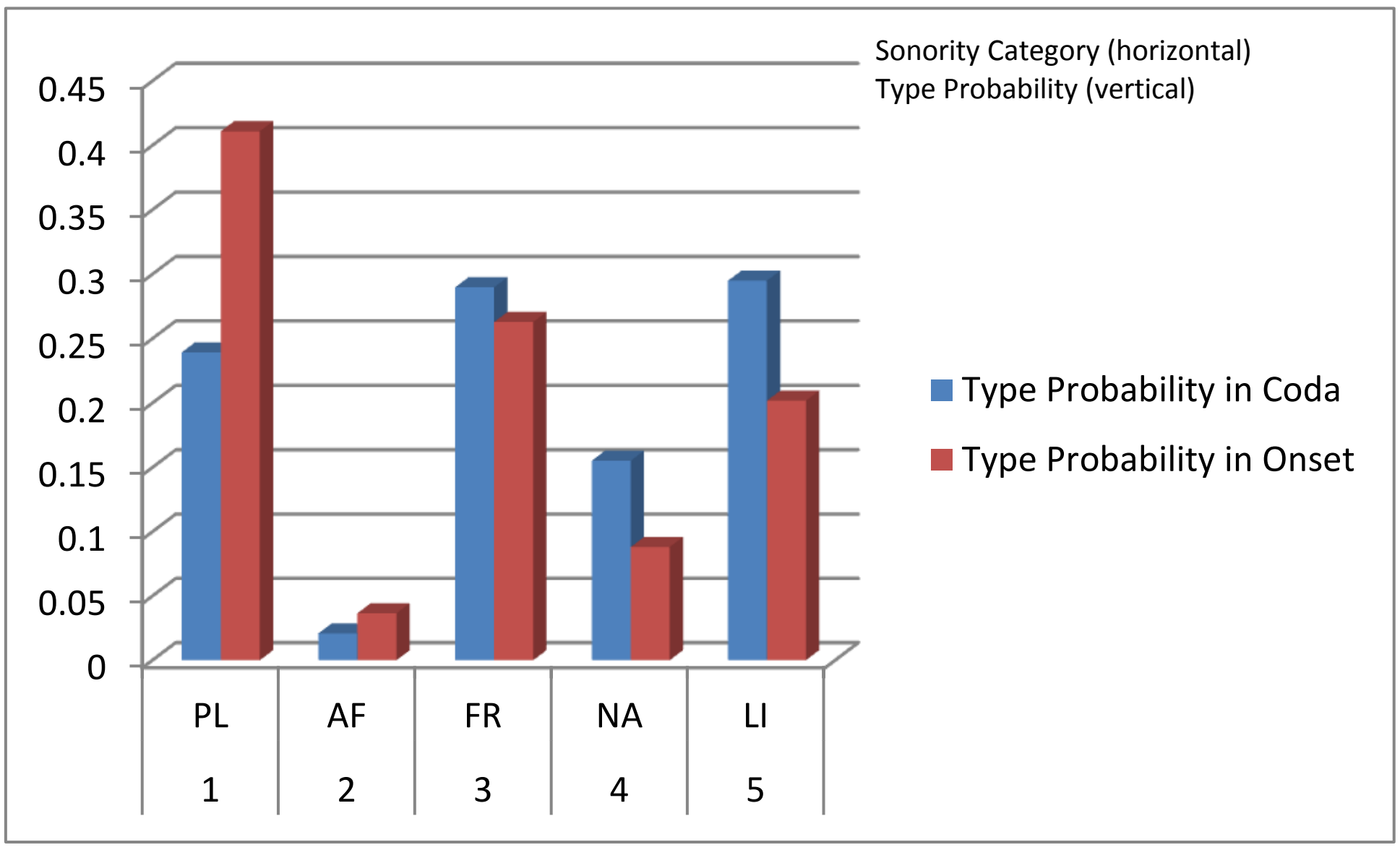

**Diagram 2 – Distribution of sonority categories in coda and onset position of a syllable contact pair in lexicon**

In chart 3 corpus frequency is shown based on the slope of syllable contacts. Considering chart 1, we face token frequency increase in chart 3 as well, while sonority slope falls from +4 to -4 (from rising slopes to falling slopes), but this increase is very mild and low. As a matter of fact the more unmarked is the syllable contact, the more frequent it shall be. But frequency increase has a very milder slope comparing with type frequency slope shown in chart 1.

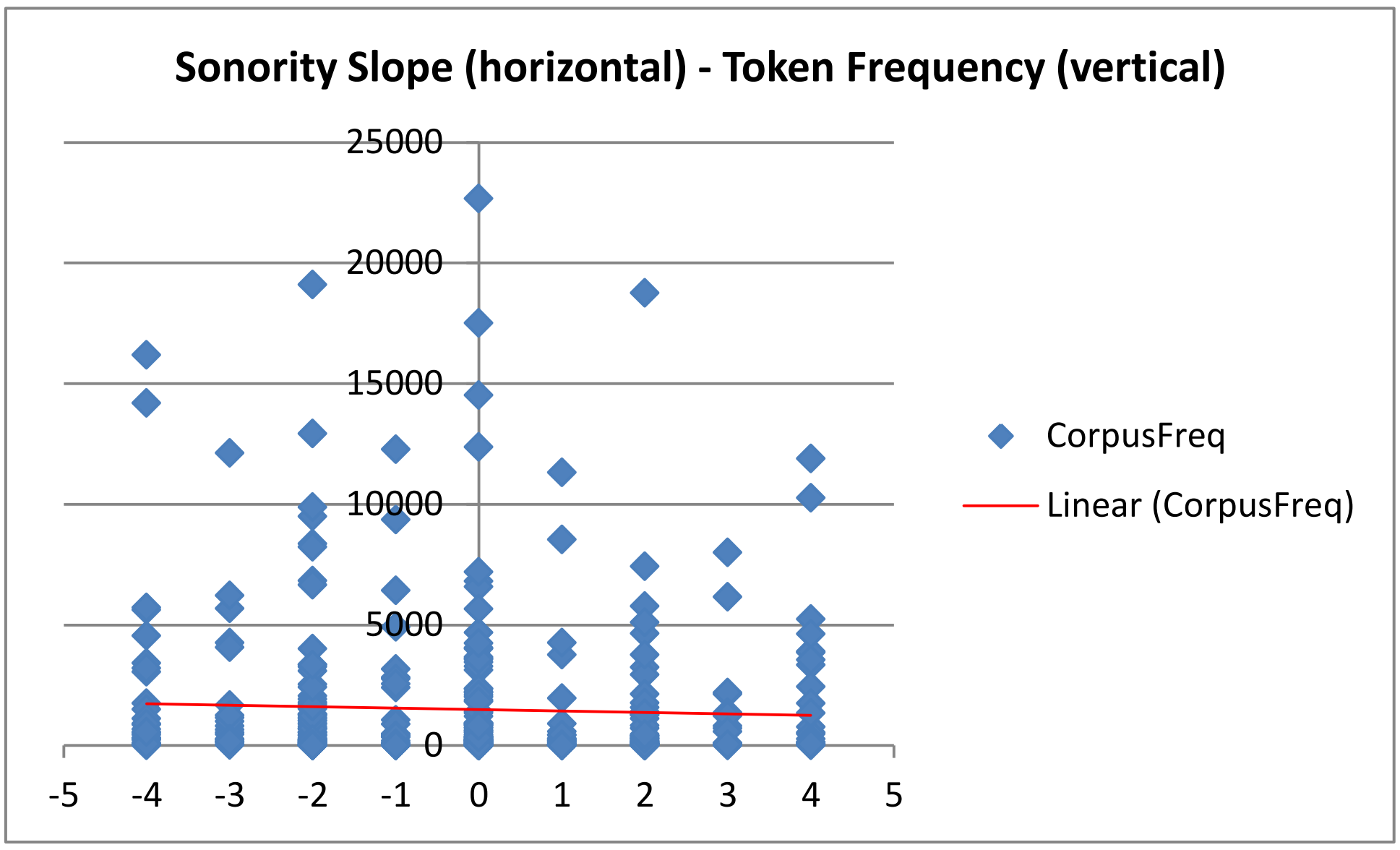

**Diagram 3 – Token frequency of sonority slopes in syllable contact pairs**

In chart 4, the probability of occurrence for all sonority categories in both coda and onset positions is compared using token frequency. The difference between this chart and chart 2 is that in chart 2 type frequencies of syllable contacts were used to calculate probabilities, while in char 4 token frequencies (frequencies in corpus) has been used to do so.

As it can be seen occurrence probability of stop categories in corpus has been sharply reduced comparing with their probability of occurrence in lexicon. This makes the slope smoother in chart 3, because of the fact that stops are less probable to occur both in onset and coda positions. Hence median states have gained more probability to occur. In addition, increase of probability of affricates to occur in both onset and coda positions is significant in corpus.

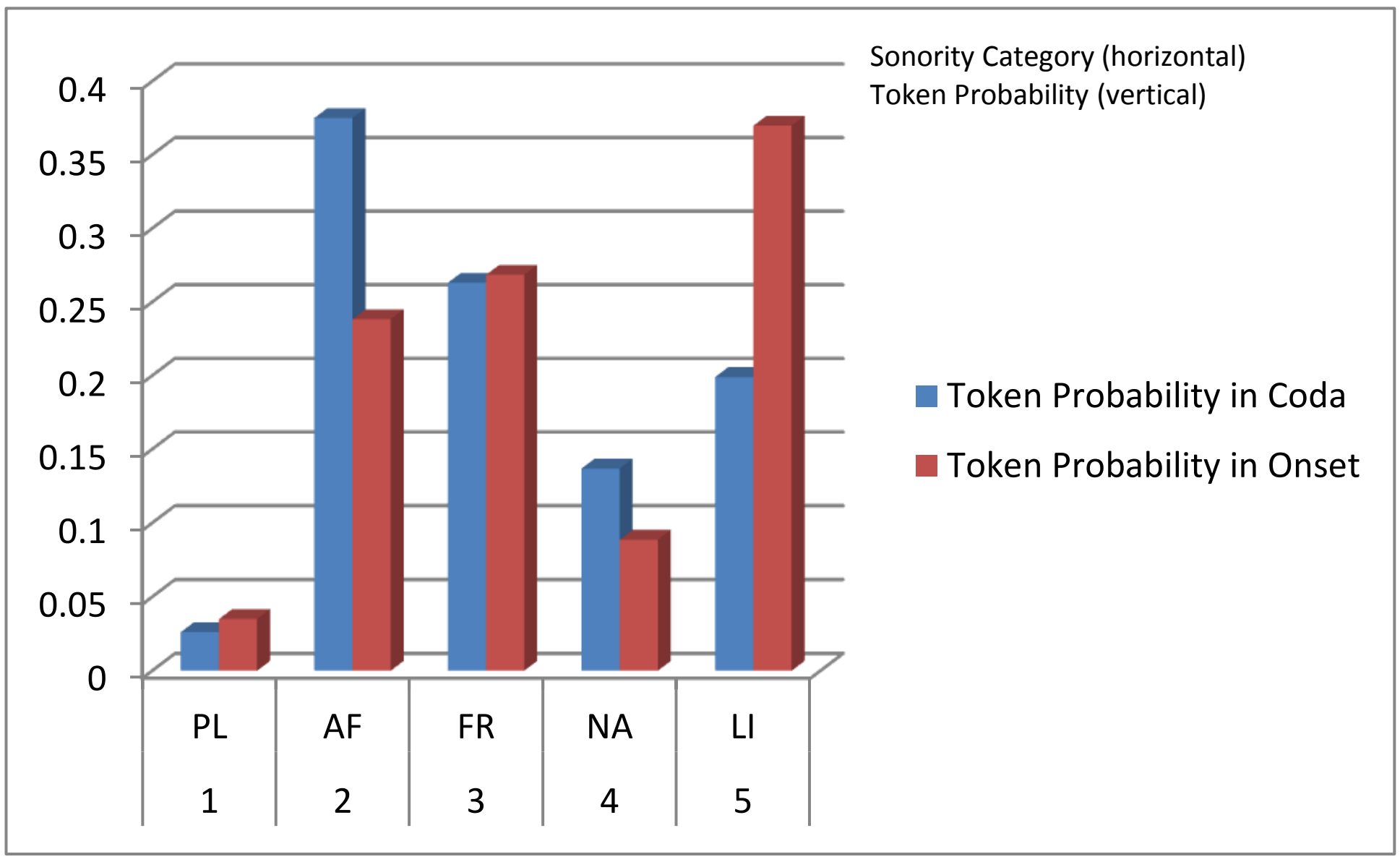

**Diagram 4 – Distribution of sonority categories in coda and onset position of syllable contact pairs in corpus**

In chart 5, PMI measure has been calculated for each syllable contact pair. For calculating the joint and distinct probabilities type frequency of the syllable contacts and their constituent consonants have been used. For each syllable contact the probability of its occurrence is divided by the probability of the first consonant in the coda position and the probability of the second consonant in the onset position of CVC.CVC sequences. The logarithm function is then applied to the result for the PMI measure to be computed. As can be seen in the chart PMI measure trend line is almost 0 everywhere. As we mentioned before, the amount of zero for Pointwise Mutual Information of two events (here the two consonants of a syllable contact) shows the tendency for the two consonants to occur independently of each other without any imposed restriction. Occurrence independence of the two consonants means there is no direct connection between type frequency and sonority slope of the syllable contacts in lexical level. This indicates evidence against the gradual frequency effects of marked and unmarked sequences according to SCL. In other words the type frequency of syllable contacts is close to their expected type frequency assuming the independence of the two consonants' occurrence. The slope in char 1 and chart 3 can be described using the tendency of less sonorant categories to occur in onset position and the tendency of sonorant categories to occur in coda position.

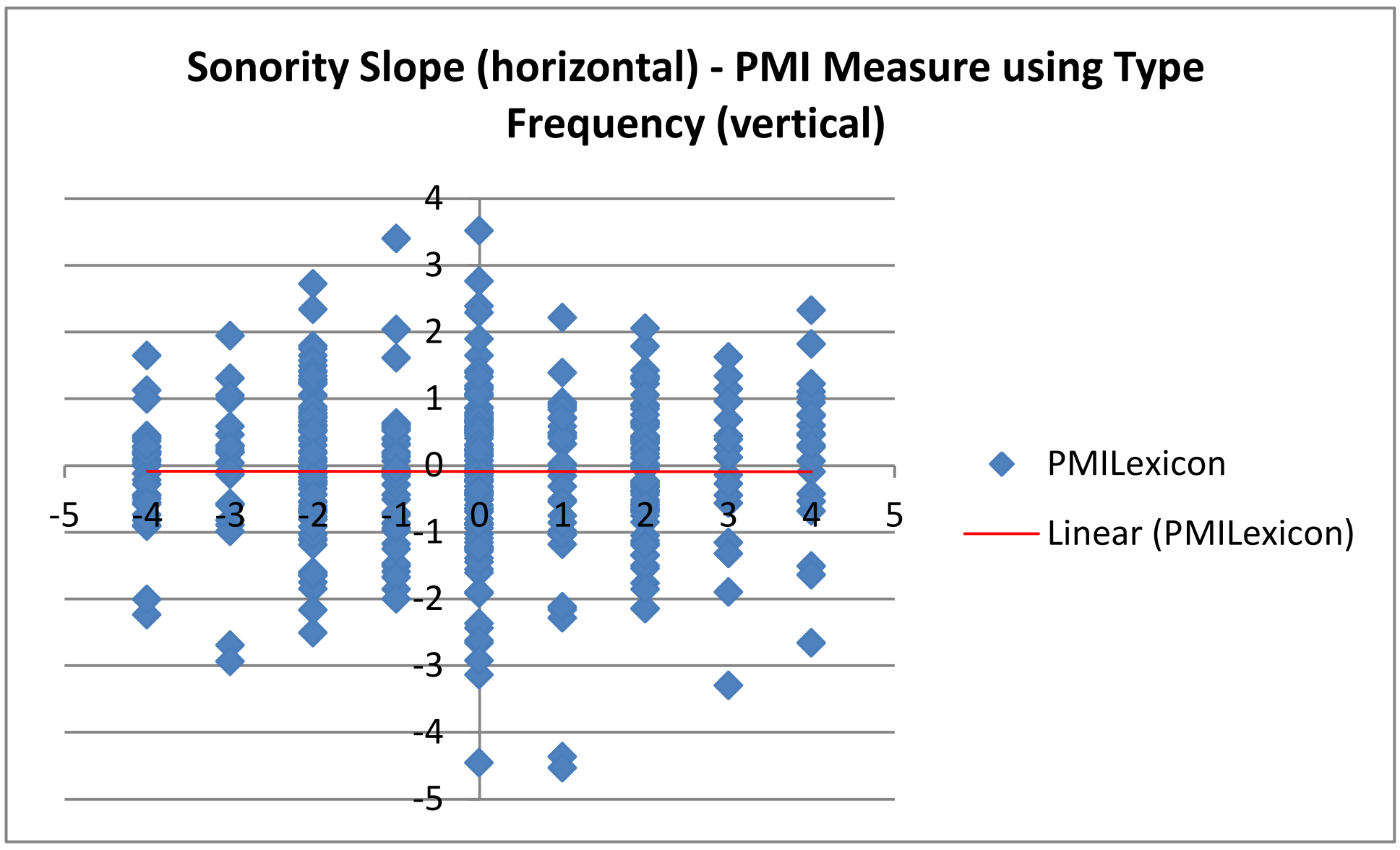

**Diagram 5 – Pointwise Mutual Information between sonority category in coda and sonority category in onset of syllable contact pairs using type frequency in lexicon**

In chart 6, PMI for occurrence of consonants in syllable contacts are calculated by the same method using token frequency instead of type frequency. The PMI measure trend line is close to zero everywhere again as in chart 5 where type frequency was used to calculate PMI. The results show that again in corpus level just like lexical level, the consonants in syllable contact occur independently and don't impose any restriction on each other.

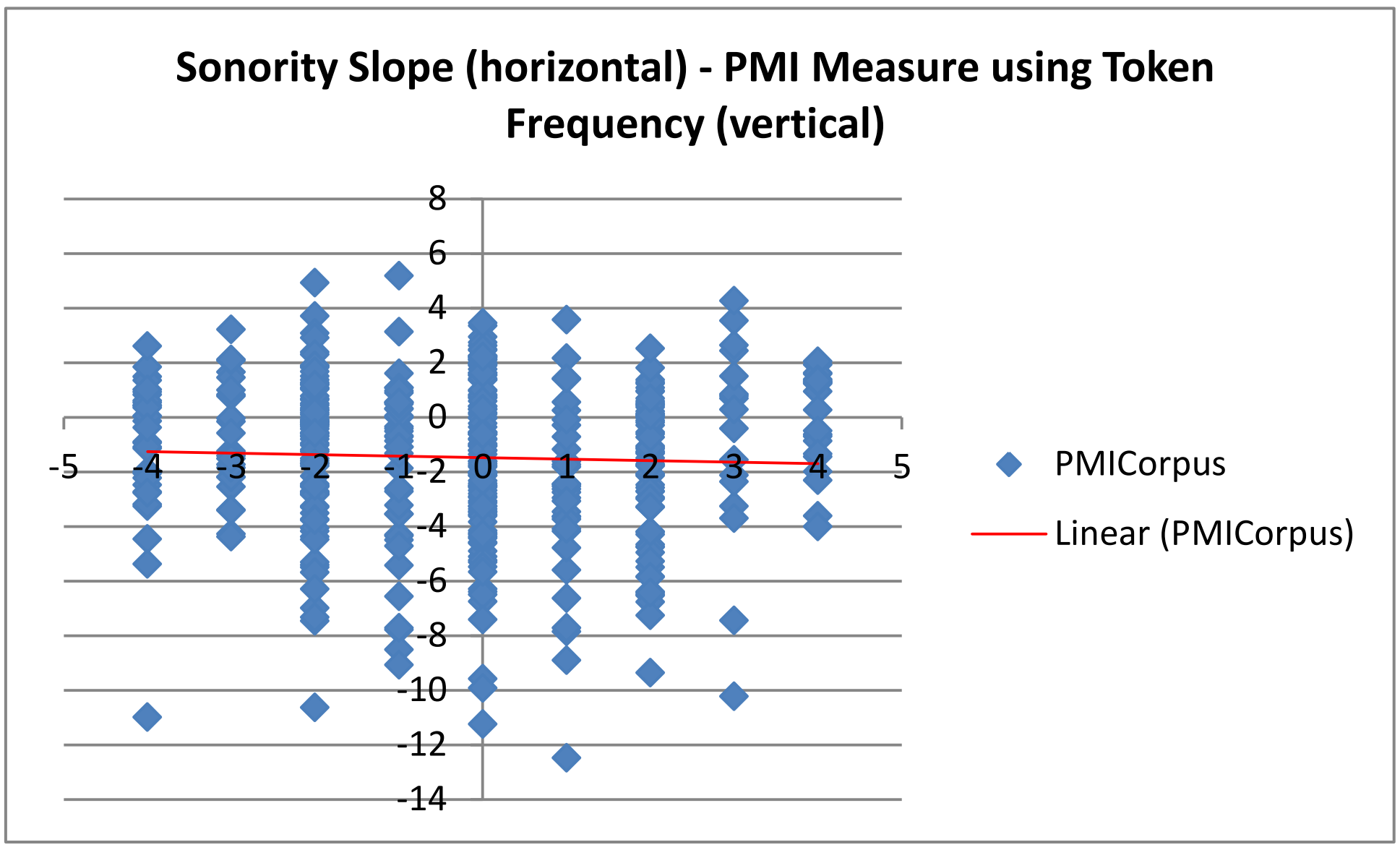

**Diagram 6 – Pointwise Mutual Information between sonority category of coda and sonority category of onset in syllable contact pairs using token frequency in corpus**

## 4. Discussion and conclusion

Results show that syllable contacts with less sonority slope are more frequent both in lexicon (chart 1) and in corpus (chart 3) but the extent to which the frequency increases from marked sequences (rising slopes) to unmarked sequences (falling slopes) is much milder in corpus.

Higher frequency of more unmarked syllable contacts (syllable contacts with falling sonority slope or less rising slope) can be the outcome of either SCL or the tendency of non-sonorant categories like stops and affricates to occur in onset position and the tendency of sonorant categories like nasals and liquids to occur in coda position. The results in chart 2 and in chart 4 provide support for the latter explanations. As shown in chart 2 stops and affricates with less sonority prefer to occur in onset position and fricatives, nasals and liquids prefer to occur in coda position so if a random syllable contact is selected it is more probable that it has a falling sonority slope. The distribution of various sonority categories in onset and coda position of syllable contacts in corpus shown in chart 5 indicates that stops' probability of occurrence is sharply decreased both in onset position and coda position. In other words sequences which use stops in their syllable contact either in onset position or coda position are less frequently used in corpus compared to other consonants. On the other hand affricates' probability of occurrence is significantly increased in corpus compared with their probability of occurrence in lexicon and with other consonants' probability of occurrence. The decrease in stops' probability of occurrence and the increase in that of affricates cause the syllable contacts' overall sonority slope to be much milder in corpus (shown in chart 3) compared to that of syllable contacts in lexicon (shown in chart 1) by reduction of boundary sonority slopes (-4, +4) and made chart3's trend line very smoother than chart 1's.

In addition, PMI measure shows that the observed frequency of syllable contacts is close to their expected frequency assuming the independence of the occurrences for the two consonants both in lexicon and in corpus and sonority slope doesn't have any impact on the frequency of syllable contacts. In other words the two consonants in a syllable contact don't impose any restriction on each other related to their sonority; Otherwise PMI would show the restrictions by a negative or positive trend in some slopes. This however doesn't mean that the consonants don't restrict each other on other dimensions. For example the consonants may impose restrictions on the place of articulation, which has not been the subject of this study, but according to sonority the overall trend in all sonority slopes for the PMI is near zero indicating overall in Persian lexicon the sonorities of the two consonants in syllable contact don't have a significant impact on each other so Syllable Contact Law is not an active phonological constraint in lexical or usage (corpus) level in Persian or it is ranked lower than FAITH constraint according to Optimality Theory framework.

If the frequency of syllable contacts with falling sonority slope is higher than syllable contacts with rising sonority slope (chart 1, chart 3), the reason is not the presence of Syllable Contact Law as a phonemic constraint; otherwise there should have been a connection between this phonemic constraint and PMI measures in charts 5 and 6. As mentioned before the higher frequency of syllable contacts with falling sonority slope is related to the tendency of non-sonorant consonants to occur in onset position and that of sonorant consonants to occur in coda position of syllable contacts.

Reluctance of stops to occur in coda position before an onset consonant and their tendency to occur in onset position before the vowels support the hypothesis that the distribution of contrasting features is closely related to the amount of perceptual salience a context provides for the contrasting features. Distinctive features of consonants (e.g. manner, place, voice) are more perceptually salient in pre-vowel contexts compared with pre-consonantal contexts (Wright, 2004). This is especially true about stops whose voicing and place of articulation features are perceptually weak in pre-consonantal contexts. According to Licensing by Cue Hypothesis (Steriade, 1997) the more feature F of segment S is perceptually salient in context C, the more S is likely to show contrasts by values of F. On the other hand if F is perceptually weak in a context, the segments won't contrast using values of F. Dispersion Theory of Contrasts (Flemming, 1995) is a similar theory explaining the role of phonetic/functional factors in phonological patterns.

The distribution of consonants in syllable contacts (charts 2, 4) supports the phonetically-based explanation for phonological patterns. Non-sonorant consonants have inherently weaker contrasting features compared with that of sonorant consonants; therefore overall they are less frequent than sonorant consonants. Furthermore, contrasting features of non-sonorant consonants and especially that of stops are more perceptually salient in pre-vocal contexts than in pre-consonantal contexts. So stops are more frequent in onset position than in coda position of syllable contacts both in lexical and usage levels (charts 2, 4).

Diachronically phonological changes cause marked structures to change into more unmarked structures. Gradually the frequency of unmarked structures gets higher in lexicon. Hence, if there is a highly ranked restriction towards language it is expected that that the frequency of structures which have not violated the restriction (unmarked structures) gets higher gradually. Based on the fact that the PMI analysis (charts 5, 6) show lack of connection between slope patterns in syllable contact and frequency (type and token), we can come into conclusion that Syllable Contact Law (SCL) as a relational constraint is not active in Persian neither as a single markedness constraint nor as a hierarchy of constraints (Gouskova, 2004; Baertsch and Davis 2005).

The results show that the diachronic frequency effect of phonological changes toward more unmarked syllable contacts is not significant, yet, phonological changes like addition, omission and change??? are active in the phonological system of Persian speakers (table 5). The interesting fact about these phonological changes is that all of them change the severe marked structures with rising sonority slope +4. Another interesting aspect of these changes is that not all the speakers of Persian use these phonological operations in day to day conversations. A sociolinguistic analysis of the social groups using these phonological changes is an interesting subject of future research. A rough guess may be that social groups whose work environment is noisy use these phonological changes more frequently to make the marked syllable contacts more perceptually robust in noise by changing them to more unmarked consonant clusters, providing richer perceptual contexts for stops.

According to the fact that in some languages like Kazakh and Kirgiz which are samples of Turkish, SCL constraints are highly ranked and make categorical distinction, the low rank of these constraints in languages like Persian must have a reason. In Persian SCL constraints not only don't have a categorical role, dividing sequences into well-formed and malformed, but also we didn't see the gradient lexical reflexes of the constraints in lexicon which was reported in English (McGowan, 2008) using PMI analysis. One of the reasons could be deterministic recognition of syllable boundary in Persian. If consonant clusters are permitted both in onset and coda or if null onsets are permitted in a language, phonological constraints will be necessary to restrict sequence of syllable contacts so that the listener can recover the boundary of syllables. For instance, consider CVCCV chain. This structure could be divided into syllabic forms using different methods; for example: (CV, CCV), (CVC, CV) and (CVCC, V), according to the fact that in Persian it is not permitted to use a null or a consonant cluster in onset position, no other restriction is required to mark the syllable boundary for listeners and the only permitted structure shall be CV,CV. In languages like English which have more complicated syllabic structure and the recognition of syllable boundary is sometimes difficult for the speakers of the language itself, some restrictions are required to minimize different states of syllable division and to simplify the recognition of syllable boundaries. The more possible states of syllable division in a language, more forceful restrictions are needed to recognize the boundary of syllables. Therefore in Persian there is no need for the application of restrictions because of its deterministic syllable boundary. Investigating the distribution of sonority in syllable contacts in some languages with deterministic syllable boundary like Persian (e.g. French) and comparing it to some languages with nondeterministic syllable boundary like English or Kazakh can provide support for this hypothesis.

## 5. References

Burquest, D. A. (1993). Phonological analysis: A functional approach. In Lingua Links library 5.0 plus. x, 314 pages. [Dallas]: SIL International Digital Resources.

Clements, G. N. (1990). The role of the sonority cycle in core syllabification. In *Papers in laboratory phonology I: Between the grammar and the physics of speech.* New York: Cambridge University Press. 283-333.

Gouskova, M. (2004). Relational hierarchies in Optimality Theory: the case of syllable contact. *Phonology* 21: pages 201-250. Cambridge Univ Press.

Harris, J. (2005). The phonology of being understood: Further arguments against sonority. *Lingua , 55*(10).

Keshavarz, M. (2000). A Sociolinguistic analysis of methathsis in Persian, *JOURNAL OF HUMANITIES,* Winter-Spring 2000; 7(1-2):16-22.

Ladefoged, Peter and Keith Johnson. 2011. *A Course in Phonetics*. $6^{th}$ edition. Wadsworth, Cengage Learning.

McGowan, K. (2008). Gradient Lexical Reflexes of Syllable Contact Law. http://kmcgowan.rice.edu/publications/mcgowan-cls45.pdf

Vennemann, T. (1988). *Preference laws for syllable structure and the explanation of sound change: With special reference to German, Germanic, Italian, and Latin.* Berlin: Mouton de Gruyter.

Selkrik, Elizabeth (1984). On the Major Class features and Syllable Theory, in M.Aronoff and R.T. Oherle, (Eds), *Language Sound Structure: Studies in Phonology Dedicated to Morris Halle by his students,* MIT Press, Cambridge, Mass.

Wright, R. A. (2004b). A review of perceptual cues and cue robustness. In B. Hayes, R. Kirchner, & D. Steriade (Eds.), *Phonetically based phonology* (pp. 34-57). Cambridge; New York: Cambridge University Press.

Steriade, D. (1997). Phonetics in phonology: the case of laryngeal neutralization. UCLA ms.

Eslami, M., et al. 2004. “vâžegân-e zâyâye zabân-e fârsi” [“Persian generative lexicon”], Proceedings of The 1st Workshop on Persian Language and Computer, May 26-27 2004, University of Tehran, with 4 the cooperation of the Research Center of Intelligent Signal Processing (RCISP), Tehran, Iran.

Ahmadkhani, M. (2010). Phonological Metathesis in Persian: Synchronic, Diachronic, and the Optimality Theory. *PAZHUHESH-E ZABANHA-YE KHAREJI* , 5-24.

Church, Kenneth W., & Patrick Hanks. 1989. Word association norms, mutual information, and lexicography. In Proceedings of the 27th. Annual Meeting of the Association for Computational Linguistics, 76–83, Vancouver, B.C. association for Computational Linguistics.

Baertsch, Karen. 1998. Onset sonority distance constraints through local conjunction. In CLS 34, Part 2: The Panels, eds. M. Catherine Gruber, Derrick Higgins, Kenneth S. Olson and Tamra Wysocki, 1-15. Chicago: Chicago Linguistic Society.

Baertsch, Karen, and Stuart Davis. 2001. Turkic C+/l/(uster) Phonology. In CLS 37: The Main Session, eds. Mary Andronis, Christopher Ball, Heidi Elston and Sylvain Neuvel. Chicago: Chicago Linguistic Society.

Baertsch, Karen. 2002. An Optimality-theoretic approach to syllable structure: The Split Margin Hierarchy. Indiana University: Ph.d. Dissertation.

Parker, Steve. 2002. Quantifying the Sonority Hierarchy. Linguistics, University of Massachusetts, Amherst: Ph.d. Dissertation